\documentclass[lettersize,journal]{IEEEtran}
\usepackage{amsmath,amsfonts}
\usepackage{algorithmic}
\usepackage{algorithm}
\usepackage{array}
\usepackage[caption=false,font=normalsize,labelfont=sf,textfont=sf]{subfig}
\usepackage{textcomp}
\usepackage{stfloats}
\usepackage{url}
\usepackage{verbatim}
\usepackage{graphicx}
\usepackage[numbers,sort&compress]{natbib}
\usepackage{threeparttable}
\usepackage{pifont}
\usepackage{multirow}
\usepackage{amssymb}
\usepackage[table]{xcolor}

\newcommand{\stwo}{Sentinel-2}
\newcommand{\datasetname}{M3R-CR}
\newcommand{\cmark}{\ding{51}}%
\newcommand{\xmark}{\ding{55}}%

\usepackage[normalem]{ulem}
\usepackage{xcolor}
\newcommand{\revise}{\color{black}}
\begin{document}

\title{Multi-Modal and Multi-Resolution Data Fusion for High-Resolution Cloud Removal: A Novel Baseline and Benchmark}

\author {    Fang~Xu,
        \and Yilei~Shi,~\IEEEmembership{Member,~IEEE,}
        \and Patrick~Ebel,
        \and Wen~Yang,~\IEEEmembership{Senior~Member,~IEEE,}
        \and Xiao~Xiang~Zhu,~\IEEEmembership{Fellow,~IEEE,}
\thanks{This work is supported by the German Federal Ministry of Education and Research (BMBF) in the framework of the international future AI lab "AI4EO -- Artificial Intelligence for Earth Observation: Reasoning, Uncertainties, Ethics and Beyond" (grant number: 01DD20001). The work of W. Yang is supported by the National Natural Science Foundation of China (NSFC) Regional Innovation and Development Joint Fund (No. U22A2010) and NSFC General Program (No. 62271355). 
\\
(\textit{Corresponding Author: Xiao Xiang Zhu})}
\thanks{Fang~Xu is with the School of Electronic Information, Wuhan University, Wuhan 430072, China, and also with the Data Science in Earth Observation, Technical University of Munich, Munich 80333, Germany (\emph{e-mail: xufang@whu.edu.cn})}
\thanks{Yilei~Shi is with the Remote Sensing Technology, Technical University of Munich, Munich 80333, Germany. \emph{e-mail: yilei.shi@tum.de}}
\thanks{Patrick~Ebel is with the Data Science in Earth Observation, Technical University of Munich, Munich 80333, Germany (\emph{e-mail: patrick.ebel@tum.de})}
\thanks{Wen~Yang is with the School of Electronic Information, Wuhan University, Wuhan 430072, China (\emph{e-mail: yangwen@whu.edu.cn})}
\thanks{Xiao~Xiang~Zhu is with the Data Science in Earth Observation, Technical University of Munich, Munich 80333, Germany and with Munich Center for Machine Learning, Munich, Germany. (\emph{e-mail: xiaoxiang.zhu@tum.de})}

}

\markboth{submitted to IEEE Transaction on Geoscience and Remote Sensing}%
{Shell \MakeLowercase{\textit{et al.}}: A Sample Article Using IEEEtran.cls for IEEE Journals}


\maketitle

\begin{abstract}
Cloud removal is a significant and challenging problem in remote sensing, and in recent years, there have been notable advancements in this area. 
However, two major issues remain hindering the development of cloud removal: 
the unavailability of high-resolution imagery for existing datasets and {\revise the absence of evaluation regarding the semantic meaningfulness of the generated structures}.
In this paper, we introduce \datasetname{}, a benchmark dataset for high-resolution Cloud Removal with Multi-Modal and Multi-Resolution data fusion. 
\datasetname{} is the first public dataset for cloud removal to feature globally sampled high-resolution optical observations, paired with radar measurements and pixel-level land cover annotations. 
With this dataset, we consider the problem of cloud removal in high-resolution optical remote sensing imagery by integrating multi-modal and multi-resolution information{\revise. In this context, we have to take into account the alignment errors caused by the multi-resolution nature, along with the more pronounced misalignment issues in high-resolution images due to inherent imaging mechanism differences and other factors.} Existing multi-modal data fusion based methods, which assume the image pairs are aligned {\revise accurately at pixel-level}, are thus not appropriate for this problem. 
To this end, we design a new baseline named Align-CR to perform the low-resolution SAR image guided high-resolution optical image cloud removal{\revise. It gradually warps and fuses the features of the multi-modal and multi-resolution data during the reconstruction process, effectively mitigating concerns associated with misalignment.}
In the experiments, we evaluate the performance of cloud removal by analyzing the quality of visually pleasing textures using image reconstruction metrics and further analyze the generation of semantically meaningful structures using a well-established semantic segmentation task.
The proposed Align-CR method {\revise is superior to other baseline methods} in both areas. 
The project is available at \url{https://gitlab.lrz.de/ai4eo/M3R-CR}.
\end{abstract}


\section{Introduction}

\IEEEPARstart{R}{emote}
sensing imagery has been receiving considerable attention as a major promising prospect in various applications such as Earth observation and environmental monitoring~\citep{xia2018dota, yuan2020deep, requena2021earthnet2021, earthnets4eo}. 
However, haze and clouds in the atmosphere affect the transmission of electromagnetic signals and lead to a deficiency of surface information~\citep{duan2020multi}, severely limiting the potential of optical remote sensing imagery~\citep{9956865}.
Cloud removal aims at reconstructing the cloud-contaminated regions to counteract the degradation caused by clouds, and thus {\revise is essential for remote sensing interpretations and applications}.
To advance the state-of-the-art in cloud removal, many datasets, such as RICE-II~\citep{lin2019remote} and SEN12MS-CR~\citep{ebel2020multisensor}, have been proposed in recent years. Most of these datasets are built based on Landsat-8 or Sentinel-2 imagery. However, remote sensing technology has given rise to the next-generation satellites, which can provide optical imagery with higher spatial resolution~\citep{2021_5b8add2a, cornebise2022open}. Such data allows geometric analysis on a finer scale while posing new challenges for recovering cloud-covered regions with the corresponding levels of detail. The community is very productive when it comes to proposing novel methods, 
while {\it it remains an open question whether and to what extent the cloud removal techniques developed on current datasets with relatively low spatial resolutions can generalize to high-resolution ones}. 
The unavailability of high-resolution cloud removal datasets severely restricts the development of cloud removal algorithms for recovering clear edges and rich texture details of high-resolution remote sensing imagery.

To compare different cloud removal algorithms, existing work almost exclusively relies on metrics such as PSNR and SSIM to evaluate the quality of reconstructed images, which can provide quantitative information about the visual quality of the reconstructed images. However, it is also essential that the reconstructed remote sensing images are suitable for semantic analysis, in addition to being visually appealing. These visual metrics do not provide a comprehensive evaluation of the reconstructed image quality, as {\it it is uncertain whether the resulting visually appealing images perform well in the subsequent semantic tasks}. To identify the usability of the reconstructed images, it is desirable to evaluate them in terms of generating semantically meaningful structures.

\begin{figure*}[!t]
\centering
\includegraphics[width=1.\linewidth]{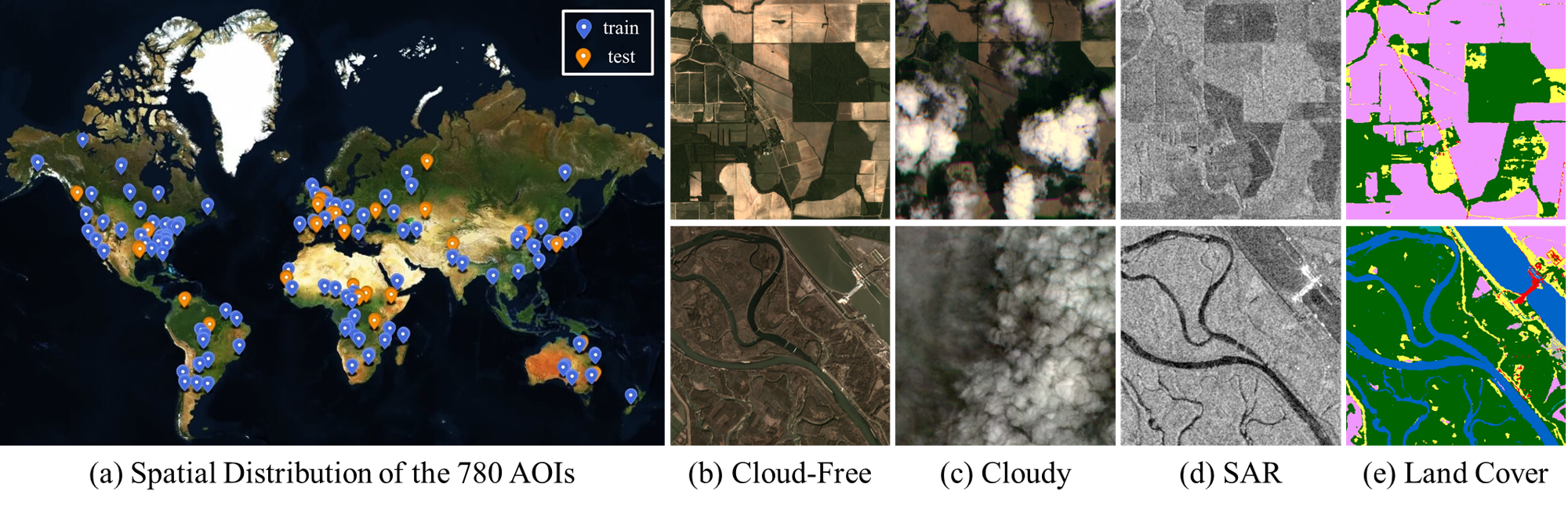}
\caption{Visualization of the \datasetname{} dataset. (a) Spatial distribution of the globally sampled 780 areas of interest. (b) Cloud-free optical observations from PlanetScope. (c) Cloudy optical observations from PlanetScope. (d) SAR observations from Sentinel-1 (visualized with the VV polarization mode). (e) Land cover maps from  WorldCover. They are scaled to the same size for better viewing. }	
\label{fig:dataset}
\vspace{-0.1cm}
\end{figure*}

In this paper, we introduce the {\it \datasetname{}} dataset, illustrated in Fig.~\ref{fig:dataset}, which consists of paired cloudy and corresponding cloud-free optical image tiles collected from Planet satellite imagery~\citep{planet} at a spatial resolution of $3m$. Compared to existing publicly available cloud removal datasets which are mostly built on Landsat-8 data with $30m$ resolution or Sentinel-2 data with $10m$ resolution, our dataset addresses the current lack of cloud removal datasets with high resolution. 
Furthermore, our dataset collects corresponding pixel-level land cover annotations from the WorldCover product~\citep{zanaga2021esa}. This provides the opportunity to validate the effectiveness of the cloud removal methods in generating globally distributed and semantically meaningful structures through a well-established remote sensing task.

Cloud removal is a highly ill-posed problem due to the loss of surface information. Previous studies have indicated that the ill-posedness can be reduced by resorting to Synthetic Aperture Radar (SAR) data, which is cloud-penetrable and inherently reflects the geometrical characteristics of ground objects~\citep{ebel2020multisensor, gao2020cloud, GLF-CR}. 
{\revise In these studies, SAR data with resolutions comparable to optical data, or downsampled higher-resolution SAR data, is often utilized as a complementary data source for cloud removal. 
However, compensating for Planet data presents further intricacies, as acquiring high-resolution SAR data with a resolution similar to that of Planet data, such as TerraSAR-X and COSMO-SkyMed satellite data, is typically unaffordable~\citep{luo2022monitoring}, especially on a global scale.
Yet, there's a promising alternative: the freely available global Sentinel-1 SAR data from the European Space Agency, albeit of a lower resolution, can be incorporated into our M3R-CR dataset to enhance the cloud removal performance.}
Compared with existing cloud removal datasets like SEN12MS-CR, which mostly explore the fusion of Sentinel-1 SAR data and Sentinel-2 optical data at a consistent resolution of 10m,
our \datasetname{} dataset helps to understand a more practical yet more complex problem: 
multi-modal and multi-resolution data fusion based cloud removal, in which {\revise the alignment inaccuracies stemming from resolution difference, more pronounced misalignment issues in high-resolution images due to factors like field-of-view mismatch and disparity, have to be accounted for}.

To address the problem of multi-modal and multi-resolution data fusion based cloud removal, we propose a novel method called Align-CR, where the low-resolution SAR images guide the reconstruction of high-resolution cloud-free images from cloudy images. Specifically, {\revise Align-CR progressively refines the alignment of feature maps from the multi-modal and multi-resolution data using the deformable convolution~\citep{dai2017deformable}, compensating for their misalignment and thereby enhancing the fusion process.} 
Based on the \datasetname{} dataset, we benchmark representative cloud removal methods and analyze their performance in generating visually pleasing textures by image reconstruction metrics and in generating semantically meaningful structures with a well-established semantic segmentation task. Extensive evaluations demonstrate that the Align-CR method achieves the best performance on the vast majority of our benchmark tests. 
In summary, our contributions are as follows:
\begin{itemize}
	\item We construct \datasetname{}, a dataset collected specially for cloud removal to advance the field. It is the public dataset with the highest spatial resolution to date and comprises a large number of regions all around the world with auxiliary SAR imagery and land cover information.
	
	\item We benchmark state-of-the-art cloud removal algorithms on the proposed \datasetname{} dataset and perform extensive evaluations of the recovered semantic information on a well-established semantic segmentation task.
	
	\item We propose a novel cloud removal algorithm, Align-CR, which can better explore complementary information across the multi-modal and multi-resolution data to better reconstruct the occluded regions.
	
\end{itemize}  

The remainder of this paper is structured as follows. 
In Sec.~\ref{sec: related_work}, we provide a review of the existing datasets and algorithms for cloud removal. 
Then, Sec.~\ref{sec: data} describes the proposed \datasetname{} dataset in detail.
Our proposed Align-CR method is introduced in Sec.~\ref{sec: methodology} and experimental results are discussed in Sec.~\ref{sec: evaluations}.
Finally, discussions for further work and conclusions are presented in sec.~\ref{sec:discussion} and sec.~\ref{sec:conclusion}, respectively.

\section{Related Work}
\label{sec: related_work}
\subsection{Datasets for Cloud Removal}
\begin{table*}[!t]
\caption{Comparison between Planet-CR and existing publicly available cloud removal datasets. LC is short for land cover map. }
\label{tab:datasets}
\centering
\begin{threeparttable}          
\resizebox{1.\textwidth}{!}{
\begin{tabular}{lccccccccccc}
\hline
\multirow{2}{*}{Dataset} && \multicolumn{5}{c}{Optical Image} 
&& \multirow{2}{*}{w/ SAR} & \multirow{2}{*}{w/ LC} \\ \cline{3-7} 
&& source       & resolution & \#AOIs &width    & \#images &&         &        \\ \hline
RICE-I~\citep{lin2019remote}          
&& Google Earth & $<15m$     & /      &512      & 500      && \xmark  & \xmark \\
RICE-II~\citep{lin2019remote}         
&& Landsat-8    & $ 30m$     & /      &512      & 450      && \xmark  & \xmark \\
STGAN~\citep{9093564}                 
&& \stwo        & $ 10m$     & 945    &256      & 3,101    && \xmark  & \xmark \\
SEN12MS-CR~\citep{ebel2020multisensor}
&& \stwo        & $ 10m$     & 169    &256      & 122,218  && \cmark  & \xmark \\
SEN12MS-CR-TS~\citep{ebel2022sen12mscrts}
&& \stwo     & $ 10m$        & 53     &256      & 15,578   && \cmark  & \xmark \\
WHUS2-CR~\citep{li2021deep}           
&& \stwo        & $ 10m$     & 36     &256      & 17,182   && \xmark  & \xmark \\
Scotland\&India$^*$~\citep{rs14061342}
&& \stwo        & $ 10m$     & 445    &256      & 445      && \cmark  & \xmark \\\hline
\datasetname{}                             
&& PlanetScope  & $  3m$     & 780    &300      & 63,000   && \cmark  & \cmark \\\hline
\end{tabular}
}
\begin{tablenotes}   
\small             
\item[*] The cloudy observations in the dataset are synthesized.    
\end{tablenotes}           
\end{threeparttable}   
\end{table*}

To promote the progress of deep learning-based cloud removal, several datasets for cloud removal have been proposed. We provide an overview of publicly available cloud removal datasets, as shown in Tab.~\ref{tab:datasets}. 
RICE-I~\citep{lin2019remote} contains 500 pairs of cloudy and cloud-free images collected on Google Earth by setting whether to display the cloud layer, which only contains filmy, partly-transparent clouds. RICE-II~\citep{lin2019remote} contains 450 paired images derived from the Landsat-8 OLI/TIRS dataset, where the acquisition time of cloudy and cloud-free images at the same location is less than 15 days. However, the size of both datasets is relatively small, and the data in them is geographically and topographically homogeneous. 
{\revise STGAN~\citep{9093564} comprises approximately 3,000 images of $256\times256$ pixels, obtained from 945 distinct tiles worldwide, where each tile is a square image from the Sentinel-2 satellite with a size of  $10980\times10980$ pixels.} The cloud cover for each cloudy image is between 10\% and 30\%. It excludes images with insufficient visible ground upon manual inspection. Then SEN12MS-CR~\citep{ebel2020multisensor} collects around $150,000$ samples, containing different types of real-life clouds, from $169$ non-overlapping areas of interest (AOI) sampled across all inhabited continents during all meteorological seasons. Each AOI is composed of a pair of orthorectified, geo-referenced Sentinel-2 images, with one image being cloudy and the other being cloud-free. To develop cloud removal models that are robust to extensive cloud coverage conditions, each AOI includes an additional co-registered Sentinel-1 SAR image. The related SEN12MS-CR-TS dataset~\citep{ebel2022sen12mscrts} is structured likewise while featuring multi-season repeated measurements.
However, a common limitation of the aforementioned datasets is that they are not stratified by land cover types, making it impossible to assess the generalizability of cloud removal methods over the individual land cover types that may be of interest to specific applications like vegetation monitoring~\citep{ROGAN2002143, lima2019comparing, ngadze2020exploring} and water resources monitoring~\citep{FISHER2016167}. 
To evaluate the effectiveness of cloud removal methods over different land cover types, WHU2-CR~\citep{li2021deep} selects $36$ locations from all over the world according to three main land covers: urban, vegetation, and bare land, and produces about $20,000$ pairs of cloudy and cloud-free Sentinel-2 images. Czerkawski et al.~\citep{rs14061342} construct a cloud removal dataset, especially in the context of crop monitoring. The dataset contains paired Sentinel-1 and Sentinel-2 Images for 2 locations in Scotland and India. Notably, the cloudy observations in the dataset are synthesized {\revise by overlaying cloud-free images with artificial cloud masks. To further simulate real-world conditions, plausible cloud masks with a coverage area between 10\% and 50\%, acquired from real images, are also incorporated.} 
{\revise Although simulated cloudy observations provide certain benefits, previous research has indicated that models trained on synthesized data poorly generalize to the scenario of real cloud-covered satellite imagery and their spectral characteristics~\citep{ebel2020multisensor}}.

In addition to the previously mentioned public datasets, there are also several non-public datasets. For example, Enomoto et al.~\citep{enomoto2017filmy} create a dataset by combining {\revise clouds simulated by Perlin noise with eight comparatively cloudless WorldView-2 images to generate obscured images for learning}. Cresson et al.~\citep{cresson2019optical} use Sentinel-2 images acquired over the province of Tuy, Burkina Faso to construct experiments. Gao et al.~\citep{gao2020cloud} collect two simulated datasets with around $35\%$ cloud cover based on Gaofen-2 optical imagery and airborne optical imagery, respectively, and a real dataset based on Sentinel-2 optical imagery. These datasets are relatively small in size and contain a limited number of scenarios. While featuring high-resolution imagery, cloudy observations are all simulated. 

The majority of datasets containing real-life clouds are based on medium-resolution Sentinel-2 data. It is hard to develop and evaluate the removal of clouds in high-resolution imagery with salient structures and abundant textured features.
Our \datasetname{} dataset aims to advance the task of cloud removal from high-resolution imagery by releasing the cloudy and cloud-free PlanetScope data in combination with Sentinel-1 SAR data and WorldCover land cover maps. 
The inclusion of Sentinel-1 SAR data enables the exploration of multi-modal fusion for cloud removal, and the inclusion of WorldCover land cover product enables disentangling the performance over different land cover types and evaluating the quality of recovered semantic information.
It is worth noticing that Sarukkai et al.~\citep{9093564} train a baseline land classification model and evaluate its accuracy from cloudy, cloud-free, and predicted cloud-free images, respectively, to demonstrate the power of the predicted cloud-free images for downstream use. However, cloud removal is a pixel-level reconstruction task. Hence, an image-level classification task cannot adequately reflect the quality of predicted cloud-free images, as the correct class can be inferred even if the image is partially occluded~\citep{gu2022explicit}. Within our \datasetname{} dataset, a pixel-level classification task, i.e., semantic segmentation, is provided---which is more suitable for evaluating the power of cloud removal methods on detailed semantic information recovery.

\subsection{Algorithms for Cloud Removal}
Cloud removal in optical remote sensing imagery is a long-standing research problem.
Most early developments to addressing this problem involve utilizing spatial correlation or frequency difference between cloudy and non-cloudy areas~\citep{liu1984new, zhang2007gaps, helmer2005cloud}. 
For example, Zhu et al.~\citep{zhu2011modified} utilize a modified neighborhood similar pixel interpolator approach to remove the thick clouds in Landsat images by predicting the value of cloud-contaminated pixels from neighboring similar pixels.  
Shen et al.~\citep{shen2014effective} execute a homomorphic filter in the frequency domain to remove thin clouds in the Landsat and GaoFen-1 data by suppressing the low-frequency information while enhancing the high-frequency information.  
Currently, deep learning-based methods are gaining considerable attention. They have the potential to solve many of the problems that arise in traditional cloud removal methods and achieve impressive results~\citep{sun2019cloud, li2020thin, liu2021sactnet}. 
For example,  Multispectral conditional Generative Adversarial Networks (McGANs), leveraging the remarkable generative capabilities of conditional Generative Adversarial Networks (cGANs), remove simulated clouds from Worldview-2 imagery by extending the input channels of cGANs to be compatible with multispectral input~\citep{enomoto2017filmy}. Cloud-GAN learns the mapping between cloudy and cloud-free Sentinel-2 imagery using a cyclic consistent generative adversarial network~\citep{singh2018cloud}. RSC-Net estimates the cloud-free output with the contaminated Landsat-8 imagery based on an encoding-decoding framework consisting of multiple residual convolutional layers and residual deconvolutional layers~\citep{li2019thin}. 

As the clouds thicken and the cloud-covered region is dominant, the challenge of removing the clouds in optical remote sensing imagery intensifies.
In such cases, it is essential to employ auxiliary images as reference data to ascertain the ground surface information obscured by the clouds. 
A  popular alternative method is to utilize multi-temporal images to address this issue, i.e., estimating the missing information by integrating cloud-free correspondence images acquired at different time~\citep{darbaghshahi2021cloud, sebastianelli2022plfm, chen2019thick}. 
For example, Sarukkai et al.~\citep{9093564} capture the correlations across multi-temporal cloudy images over an area by using a spatio-temporal generator network to approximate a cloud-free Sentinel-2 image.
Czerkawski et al.~\citep{rs14061342} use the historical cloud-free optical data as a source of prior information to inpaint the cloud-affected regions in Sentinel-2 images. 
However, these methods may be less effective in areas with frequent cloud cover, and may not be applicable to time-critical applications. When encountering continual cloudy days~\citep{6422379}, cloud-free reference data from an adjacent period is largely unavailable~\citep{GLF-CR}, resulting in the same issue of insufficient ground surface information. 

Therefore, a series of works explore the potential of using SAR data, {\revise which can penetrate cloud cover to acquire ground information beneath the clouds}, as auxiliary data for cloud removal in optical imagery~\citep{li2023hs2p, han2023former, li2019thick}. For example, DSen2-CR~\citep{meraner2020cloud} uses a deep residual neural network to predict the target cloud-free optical image from the concatenation of Sentinel-1 SAR image and Sentinel-2 optical image. Simulation-Fusion GAN~\citep{gao2020cloud} fuses SAR and corrupted optical imagery with two generative adversarial networks to acquire the cloud-free results of simulated cloudy Gaofen-2 data and real cloudy Sentinel-2 data. GLF-CR~\citep{GLF-CR} incorporates the contribution of Sentinel-1 SAR image in restoring reliable texture details and maintaining global consistency to reconstruct the occluded region of Sentinel-2 optical image. 
However, these methods are developed on data with relatively low spatial resolutions. 
The problem of cloud removal in high-resolution imagery remains largely under-explored.

\section{Data}
\label{sec: data}
\subsection{Curation of \datasetname{}}
\label{sec:curation}
To develop cloud removal methods that are equally applicable to heterogeneous {\revise Earth} observation and evaluate the generalisability, it is desirable to curate highly representative locations. 
To this end, we sample the geospatial locations of 780 non-overlapping AOIs that are distributed over all continents and meteorological seasons of the globe, as shown in Fig.~\ref{fig:dataset}(a). 
Each AOI is composed of a quartet of orthorectified, geo-referenced cloudy, and cloud-free optical images, as well as the corresponding SAR image and land cover map.
As it is not possible to capture both cloudy and cloud-free views of a particular location simultaneously, we can only collect the cloud-free observations that are temporally close to the original cloudy observations as the reference. 
Although certain factors such as sunlight condition, acquisition geometry, humidity, and change of landscape, inevitably introduce nuisances in the data, these nuisances can be considered largely negligible for a relatively large-scale split that is globally and seasonally sampled without any bias to specific conditions. {\revise Since SAR is always capable of obtaining surface information regardless of the presence of clouds}, we consistently select the SAR image with the time interval closest to that of the cloudy image.

\noindent{\bf Optical data}. High-resolution remote sensing imagery with extensive spatial-temporal coverage is not easily available. Some satellites offer very high-resolution imagery only from specific locations, which makes it impractical to construct datasets with a high diversity of globally distributed sites. 
By trading off the spatial-temporal coverage and the resolution of satellite imagery,  we collect data from Planet satellite imagery with the objective of investigating high-resolution cloud removal techniques.
Planet provides global daily data with a spatial resolution of $3m$. 
On the one hand, it allows the acquisition of paired cloudy and cloud-free images with a very short temporal offset. It could minimize the surface changes that may appear between the acquisition of cloudy and cloud-free images, thereby reducing the nuisances between cloudy reference images and cloud-free target images.
On the other hand, it allows the acquisition of heterogeneous {\revise Earth} observation data to encourage general-purpose cloud removal but not along narrowly-defined and geo-spatially distinct regions of interest. 

Specifically, we gather the cloudy and cloud-free observations from the PlanetScope Level-3B top-of-atmosphere reflectance product, which are radiometrically-, sensor-, and geometrically-corrected. It includes four spectral bands, Blue, Red, Green, and Near-Infrared. In our dataset, the average time interval between paired cloudy and cloud-free data is $2.8$ days. This is shorter than the intra-mission revisiting period of Landsat-8 and Sentinel-2 satellites and minimizes the amount of land cover change occurring between our paired data. The cloudy data features a diversity of clouds, ranging from semi-transparent to dense, and from light to heavy cloud covering.  In addition to the cloudy and reference cloud-free data, we also collect the cloud masks associated with cloudy PlanetScope images according to the Unusable Data Mask (UDM) assets provided by Planet Labs. UDM masks give information about which pixels in the images are clear or cloudy, permitting a statistical evaluation of cloud coverage and enabling cloud mask-guided methods to work.

\noindent{\bf SAR data}. Benefiting from the free and open data policy of the Copernicus program, the SAR data of our dataset originates from the Sentinel-1 mission operated by the European Space Agency. We gather the Sentinel-1 data under ascending/descending orbits with a spatial resolution of $10m$ from the Level-1 Ground Range Detected (GRD) product archived by Google Earth Engine~\citep{gorelick2017google}. The measurements are acquired in interferometric wide swath (IW) mode with two polarization channels VV (vertical transmit/vertical receive) and VH (vertical transmit/horizontal receive), which are pre-processed with thermal noise removal, radiometric calibration, terrain correction, and are converted to backscatter coefficients ($\sigma^{\circ}$) in units of decibels (dB).  {\revise In our dataset, the average time interval between the cloudy data and its corresponding SAR data is $2.2$ days.} 

\noindent{\bf Land cover map}. The land cover maps of our dataset originate from WorldCover, which is an open-access global land cover product at a $10m$ resolution released by the European Space Agency. The product achieves an overall accuracy of about 75\% across 11 land cover classes. We access the data through Google Earth Engine. In this study, {\revise we employ a simplification process to reduce complexity and noise within our analysis. We consolidate similar land cover types that might have minor differences within the original comprehensive classification. The data is reclassified into broader categories} of 6 basic land cover types according to DeepGlobe 2018~\citep{demir2018deepglobe}: forest land, rangeland, agriculture land, urban land, barren land, and water{\revise, which allows us to focus on the fundamental characteristics of each land cover type without being overwhelmed by finer distinctions}.

\begin{figure}[!t]
\centering
\subfloat[]{\includegraphics[width=0.5\linewidth]{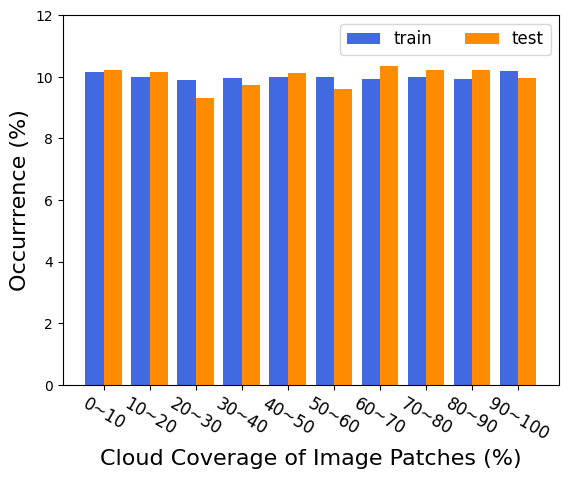}%
\label{fig:statistics-cloud}}
\hfil
\subfloat[]{\includegraphics[width=0.5\linewidth]{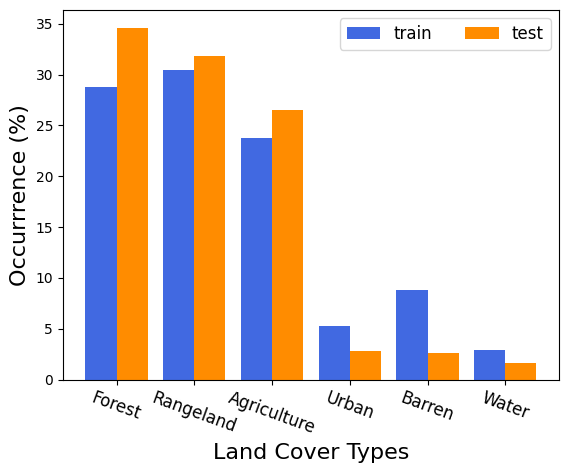}%
\label{fig:statistics-landcover}}
\caption{Statistics of the \datasetname{} dataset. (a) Distribution of cloud coverage. (b) Distribution of land cover types.}
\label{fig:statistics}
\end{figure}

We partition the dataset into train and test splits to allow for direct comparison with future works. A total of 780 AOIs are split into 660 scenes for training and 120 for testing following a random global distribution. To train the deep learning-based methods, we crop each AOI into small patches using slide windows of sensor-specific sizes. Since the spatial resolution of PlanetScope imagery is $3m$, and the spatial resolution of Sentinel-1 imagery as well as WorldCover land cover maps is $10m$, we set the corresponding slide window sizes to 300 × 300 and 90 × 90 pixels. 
We manually exclude the samples with inaccurate land cover annotations and try to distribute the samples uniformly over different cloud coverage levels as well as land cover types. We finally select 60, 000 quartets of training samples and 3,000 quartets of testing samples. The statistics about cloud coverage and land cover types can be seen in Fig.~\ref{fig:statistics}.

\subsection{Properties of \datasetname{}}
\noindent{\revise \bf Filling High-Resolution Dataset Gap}. 
Unlike existing publicly available cloud removal datasets, which are primarily built on medium-resolution Landsat-8 or Sentinel-2 data, our dataset is constructed using high-resolution PlanetScope data. As a result, it fulfills the requirement of high-resolution cloud removal datasets. It provides the opportunity to push the frontier of current cloud removal models on recovering clear edge and rich texture detail of high-resolution remote sensing imagery.

\noindent{\revise \bf Integrating Multi-Modal Multi-Resolution Data}.  
The inclusion of Sentinel-1 SAR data can provide auxiliary information to promote cloud removal performance. 
Compared with the existing cloud removal datasets like SEN12MS-CR, which mostly explore the fusion of Sentinel-1 SAR data and Sentinel-2 optical data with the same resolution of $10m$, our \datasetname{} dataset helps to understand a more practical yet more complex problem, i.e., multi-modal and multi-resolution data fusion based cloud removal.  

\noindent{\revise \bf Inclusion of Land Cover Information}. The inclusion of land cover can disentangle the performance of cloud removal methods over different land cover types on the one hand and encourages to design of a pixel-level classification task to evaluate the power of the cloud removal method in generating semantically meaningful structures on the other hand.

\section{Methodology}
\label{sec: methodology}
\subsection{Problem Statement}
Given a cloudy image $\mathcal{Y}$, the task of cloud removal aims to reconstruct a clear image $\mathcal{F}$ revealing the complete information content of the ground scene so that subsequent analysis can be reliably performed. 
The basic strategy is to deal with the cloud contamination in a single image without additional information, i.e., the problem of {\bf single image Cloud Removal} (CR), which can be formulated as:
\begin{equation}
\mathcal{F} = \textnormal{CR}(\mathcal{Y}).
\label{eq:CR}
\end{equation}
It is a highly ill-posed problem, usually solved under the assumption that the cloud-contaminated regions have similar spectral/geometrical characteristics to the remaining parts of the image. However, when it comes to the areas with high-frequency texture or different land cover types, the reconstruction performance cannot be guaranteed. Many studies resort to SAR images that are cloud-penetrable and inherently reflect the geometrical shapes of ground objects as an a priori assumption to reduce the ill-posedness. Thus, the problem of {\bf Multi-Modal data Fusion based Cloud Removal} (MMF-CR) is introduced.
It restores the clear image from a cloudy image and the corresponding SAR image $\mathcal{S}$, formulated as:
\begin{equation}
\mathcal{F} = \textnormal{MMF-CR}(\mathcal{Y}_\Omega, \mathcal{S}_\Omega),
\label{eq:MMF-CR}
\end{equation}
where $\Omega$ indicates the same spatial domain shared by the cloudy image and the corresponding SAR image. The SAR image guides most MMF-CR methods to restore the cloud-free image from pixel-to-pixel aligned cloudy and SAR images. 
However, the geo-referenced cloudy images captured by Planet satellites and SAR images captured by Sentinel-1 satellites in the \datasetname{} dataset do not meet the assumption of pixel-to-pixel alignment{\revise. This lack of alignment primarily stems from two factors.
Firstly, the non-alignment issue arises due to the resolution difference between the Sentinel-1 and Planet images. Even when the Sentinel-1 image is upsampled to match the resolution of the Planet image, the process of upsampling introduces uncertainties that contribute to the misalignment. Secondly, the misalignment issues arising from the inherent differences in imaging mechanisms and other factors, are exacerbated when it comes to higher resolution imagery. 
In essence, unlike the almost negligible misalignment in the case of Sentinel-1 and Sentinel-2 images with a consistent resolution of 10m in the SEN12MS-CR dataset, our dataset necessitates addressing the misalignment between the Sentinel-1 and Planet images.} The problem of {\bf Multi-Modal and multi-Resolution data Fusion based Cloud Removal}(MMRF-CR) is thus introduced:
\begin{equation}
\mathcal{F} = \textnormal{MMRF-CR}(\mathcal{Y}_{\Omega_y}, \mathcal{S}_{\Omega_s}|\mathcal{M})
\label{eq:MMRF-CR}
\end{equation}
where $\Omega_y$ and $\Omega_s$ respectively denote the spatial domain of the cloudy image and the SAR image, and $\mathcal{M}:\Omega_s\to\Omega_y$ denotes the pixel-level correspondence mapping operator. 
To solve the problem of MMRF-CR, it is required to precisely align the cloudy and SAR images for the cloud removal process.

\subsection{Align-CR Network}
\begin{figure*}[!t]
\centering
\includegraphics[width=1.\linewidth]{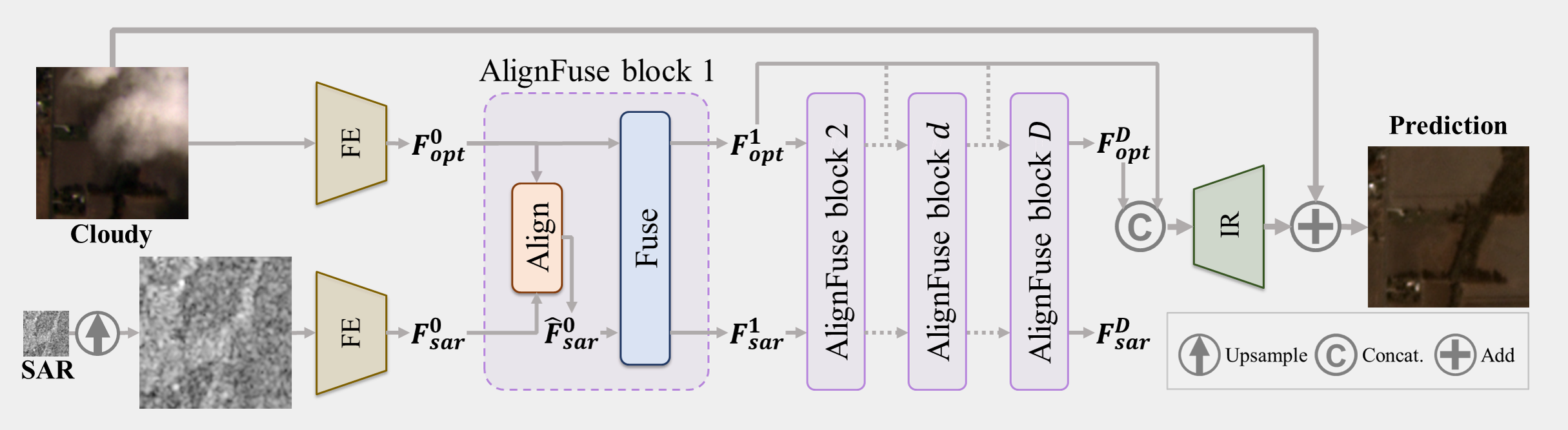}
\caption{Overview of the proposed Align-CR method. An upsampling operator is first employed to map the SAR image to the same resolution as the optical image. Then, the cloudy optical image and the upsampled SAR image are passed through their respective feature extraction (FE) blocks to extract modality-specific features. After that, the features are fed into $D$ AlignFuse blocks to obtain knowledgeable features with comprehensive information. The AlignFuse block performs alignment and fusion sequentially in the feature space. Finally, the outputs of all AlignFuse blocks are concatenated and fed to the image reconstruction (IR) block to restore the high-quality cloud-free image.}	
\label{fig:framework}
\end{figure*}
The proposed network, called Align-CR, adopts a two-stream architecture to compensate for the missing information in cloudy regions using ancillary SAR image, as shown in Fig.~\ref{fig:framework}. {\revise Rather than directly generating the entire cloud-free image from scratch, Align-CR focuses on learning the residual information, i.e., the missing or distorted features caused by the presence of clouds. 
This is accomplished by incorporating a long additive skip connection for the original cloudy image before the final output, and empirical studies have shown that this approach improves the network's convergence and enhances the quality of the final reconstructed image~\citep{lim2017enhanced, zhang2020residual, yang2022deep}.} Since the SAR image has a lower resolution, an upsampling operator is first employed to map it to the same resolution as the optical image{\revise, facilitating the generation of feature maps with consistent resolutions for subsequent processing. It is worth noting that even though both images are brought to the same resolution, non-strict alignment issues mentioned earlier, continue to persist. Subsequently}, the cloudy optical image and the upsampled SAR image are passed through their respective feature extraction blocks to extract modality-specific features $F_{opt}^{0}$ and $F_{sar}^{0}$. After that, $F_{opt}^{0}$ and $F_{sar}^{0}$ are fed into $D$ AlignFuse blocks to obtain knowledgeable features with {\revise a holistic understanding of the scene}. The AlignFuse block performs alignment and fusion sequentially in the feature space. 
\begin{gather}
\hat{F}_{sar}^{i} = H_{\textnormal{Align}}(F_{opt}^{i}, F_{sar}^{i}), \\
F_{opt}^{i+1}, F_{sar}^{i+1} = H_{\textnormal{fuse}}(F_{opt}^{i}, \hat{F}_{sar}^{i}),
\label{eq:alignfuse}
\end{gather}
where $H_{\textnormal{Align}}(\cdot)$ and $H_{\textnormal{fuse}}(\cdot)$ denote the functions of the alignment block and the fusion block, respectively{\revise. The alignment block warps the SAR feature to match the layout of the optical feature, encouraging spatial alignment of corresponding geographic features in both the optical and SAR modalities. The fusion block refines the feature maps from both modalities in an interactive manner. It dynamically transfers complementary information between the aligned feature maps for mutual enhancement. The details of these two blocks are shown in Sec.~\ref{sec:align} and Sec.~\ref{sec:fuse}.} Finally, all intermediate features $\{F_{opt}^{i}\}_{i=1}^{D}$ are aggregated to reconstruct the high-quality cloud-free image.
\subsubsection{Alignment Block}
\label{sec:align}
Ideally, the task of MMRF-CR can be considered as a two-step task that aligns the multi-modal and multi-resolution data first and reconstructs the clear image following Eq.~\ref{eq:MMF-CR} later. However, explicit pixel-to-pixel alignment to ensure that the same pixels in the SAR image and the cloudy image reflect the same ground target is very hard to achieve in our case. On the one hand, sharp edges in a high-resolution optical image cannot be exactly aligned with blurry edges in a low-resolution SAR image. On the other hand, {\revise On the other hand, due to the inherent differences in the imaging mechanisms of optical and SAR sensors, it is very difficult to resolve the unalignment caused by geometric distortions and the like}. Moreover, the occlusions in the cloudy images complicate the pixel-to-pixel alignment. Therefore, we implicitly learn the alignment process for cloud-free image reconstruction. In this paper, we resort to deformable convolution to align the SAR features to the optical features. Given two features to be aligned as input, i.e., $F_{opt}^{i}$ and $F_{sar}^{i}$, $i=0,\cdots, D-1$, The core to solve feature misalignment is to predict the offset with an offset prediction module,
\begin{equation}
	\vartriangle\!p^{i} = H_{\textnormal{offset}}(F_{opt}^{i}, F_{sar}^{i}),
	\label{eq:offset}
\end{equation}
where $H_{\textnormal{offset}}$ denotes the function of the offset prediction module, which can be implemented by general convolutional layers. With the predicted offset, the SAR feature can be warped to the optical feature using the deformable convolution (DConv),
\begin{equation}
	\hat{F}_{sar}^{i} = \textnormal{DConv}(F_{sar}^{i}, \vartriangle\!p^{i}).
	\label{eq:warp}
\end{equation}
Specifically, we adopt the Pyramid, Cascading, and Deformable convolutions (PCD)~\citep{9025464} for the alignment process. It performs the alignment in a pyramid structure, i.e., aligning the features in lower scales with coarse estimations first and propagating the aligned features and learned offsets to higher scales to refine the estimations later.
Embedding it into the network, the network's ability to model transformations can be enhanced.

\subsubsection{Fusion Block}
\label{sec:fuse}
The aligned features, i.e., $F_{opt}^{i}$ and $\hat{F}_{sar}^{i}$, are then fed into the fusion block for the transfer of complementary information. Similar to our prior work~\citep{GLF-CR}, we exploit the power of SAR information from two aspects: global fusion, to guide the global interactions among all local optical windows; local fusion, to transfer the SAR feature corresponding to cloudy areas to compensate for the missing information. Specifically, each fusion block contains an adapted SAR-guided global context interaction (SGCI) block followed by an SAR-based local feature compensation (SLFC) block. To reduce the complexity of the SGCI block, instead of adding a Swin Transformer layer (STL)~\citep{liu2021swin} after each convolutional layer in the densely connected layers, we only add an STL after the convolutional layer in the local feature fusion of the residual dense block (RDB)~\citep{zhang2020residual} for cross-window feature interaction.

\subsubsection{Loss Function}
We train the Align-CR network by minimizing the difference between the output of network $\mathbf{y}$ and the cloud-free image $\hat{\textbf{y}}$ temporally close to the input cloudy image. Though the \datasetname{} has avoided the surface changes that may appear between the acquisitions of cloudy and cloud-free images as much as possible through a short time lag for data collection, there are some inevitable nuisances as described in Sec.~\ref{sec:curation}. In this paper, we use the Charbonnier loss~\citep{8100101} which can better handle outliers for training, and extra constraints on cloudy regions are applied,
\begin{equation}
	\mathcal{L} = (1+w\mathbf{M})\odot((\mathbf{y}-\hat{\textbf{y}})^2+\epsilon^2)^\alpha
	\label{eq:loss}
\end{equation}
where $\odot$ denotes Hadamard product operator, $w$ denotes the extra weight for cloudy regions, $\mathbf{M}$ denotes the cloudy mask, $\epsilon$ and $\alpha$ are constants.

\section{Evaluations}
\label{sec: evaluations}

\subsection{Experimental Settings}
\noindent{\bf Preprocessing.}
Before the PlanetScope and Sentinel-1 data are fed into the neural networks, we apply value clipping to eliminate a small number of anomalous pixels and data scaling to improve the stability of the neural networks.
We clip the values of all bands of the PlanetScope data to [0, $10,000$] and divide by $10,000$ for all bands. We clip the VV and VH polarizations of the Sentinel-1 data to values [-25, 0] and [-32.5, 0], respectively, and rescale them to the range [0, 1]. All experiments in this paper use these preprocessing steps, following previous best practices \citep{meraner2020cloud, ebel2020multisensor}.

\noindent{\bf Implementation Details.} The proposed Align-CR network is implemented using Pytorch and trained on 2 NVIDIA Geforce RTX 3090 GPUs with a batch size of 12. During training, we randomly crop the samples into $160\times160$ patches. The Adam optimizer is used and the maximum epoch of training iterations is set to 30.  The learning rate is set to $10^{-4}$ for the whole network except for the Alignment blocks where the learning rate is set to a smaller value of $10^{-5}$. The learning rates decay by 50\% every 5 epochs after the first 10 epochs. For the network architecture, the upsampling operator adopts the nearest neighbor interpolation and the number of the AlignFuse blocks $D$ is set to 6. For the loss function, $w$, $\epsilon$ and $\alpha$ is set to $5$, $10^{-3}$ and $0.45$, respectively.

\noindent{\bf Baselines}
In this paper, we compare the proposed method with 5 baseline methods on the \datasetname{} dataset with the proposed data splits, including the single image cloud removal methods, McGAN~\citep{enomoto2017filmy} and SpA GAN~\citep{pan2020cloud}, and the multi-modal data fusion based cloud removal methods, SAR-Opt-cGAN~\citep{grohnfeldt2018conditional}, DSen2-CR~\citep{meraner2020cloud} and GLF-CR~\citep{GLF-CR}. 
Since existing multi-modal data fusion based cloud removal methods require the input SAR images to be of the same spatial resolution as the input optical images, upsampling the SAR images in the \datasetname{} dataset is necessary for these algorithms to work properly. 
Here, all multi-modal data fusion based cloud removal methods utilize the SAR images upsampled by the nearest neighbor interpolation as input. 
Additionally, to determine the benefits of including auxiliary low-resolution SAR images, we train the Align-CR network without the use of the SAR images, denoted as w/o SAR. Moreover, to validate the superiority of Align-CR in integrating multi-modal and multi-resolution information, we train the Align-CR network by removing the alignment blocks in the AlignFuse blocks, denoted as w/o Align.

\begin{table*}[!t]
\setlength\tabcolsep{2pt}
\caption{Quantitative results of visual recovery quality over different land cover types.}
\label{tab:sota}
\centering
\resizebox{1.\textwidth}{!}{
\begin{tabular}{lcccccccc}
\hline
\multicolumn{9}{l}{\multirow{2}{*}{(a) {\revise results on MAE and PSNR, to quantify the reconstruction error} }}\\
\multicolumn{9}{c}{}\\\hline
\multirow{2}{*}{} & \multicolumn{6}{c}{per class MAE ($10^{-2}$) $\downarrow$} 
& \multirow{2}{*}{MAE ($10^{-2}$) $\downarrow$} & \multirow{2}{*}{PSNR (dB) $\uparrow$} \\ \cline{2-7}
&Forest &Rangeland &Agriculture &Urban &Barren &Water & & \\ \hline
McGAN~\citep{enomoto2017filmy} 
&{\revise $3.31\!\pm\!0.10$}&{\revise $4.14\!\pm\!0.03$}&{\revise $4.91\!\pm\!0.06$}
&{\revise $4.87\!\pm\!0.05$}&{\revise $4.98\!\pm\!0.09$}&{\revise $4.37\!\pm\!0.05$}
&{\revise $4.25\!\pm\!0.00$}&{\revise $26.06\!\pm\!0.36$}\\
SpA GAN~\citep{pan2020cloud}
&{\revise $3.63\!\pm\!0.29$}&{\revise $3.72\!\pm\!0.09$}&{\revise $3.95\!\pm\!0.07$}
&{\revise $4.40\!\pm\!0.07$}&{\revise $4.28\!\pm\!0.05$}&{\revise $4.74\!\pm\!0.62$}
&{\revise $3.83\!\pm\!0.15$}&{\revise $26.22\!\pm\!0.27$}\\ \hline
SAR-Opt-cGAN~\citep{grohnfeldt2018conditional}
&{\revise $3.46\!\pm\!0.46$}&{\revise $3.31\!\pm\!0.34$}&{\revise $3.49\!\pm\!0.18$}
&{\revise $3.58\!\pm\!0.05$}&{\revise $3.58\!\pm\!0.15$}&{\revise $4.61\!\pm\!0.47$}
&{\revise $3.62\!\pm\!0.39$}&{\revise $27.47\!\pm\!0.74$}\\
DSen2-CR~\citep{meraner2020cloud}
&{\revise $2.85\!\pm\!0.05$}&{\revise $2.89\!\pm\!0.03$}&{\revise $3.10\!\pm\!0.01$}
&{\revise $3.57\!\pm\!0.04$}&{\revise $3.42\!\pm\!0.01$}&{\revise $3.53\!\pm\!0.01$}
&{\revise $2.93\!\pm\!0.04$}&{\revise $28.86\!\pm\!0.08$}\\ 
GLF-CR~\citep{GLF-CR}
&{\revise $2.30\!\pm\!0.01$}&{\revise $2.56\!\pm\!0.01$}&{\revise $2.93\!\pm\!0.00$}
&{\revise $3.30\!\pm\!0.01$}&{\revise $3.22\!\pm\!0.00$}&{\revise $2.76\!\pm\!0.02$}
&{\revise $2.55\!\pm\!0.01$}&{\revise $30.00\!\pm\!0.01$}\\
\hline
Ours (wo/ SAR)    
&{\revise $2.51\!\pm\!0.10$}&{\revise $2.57\!\pm\!0.02$}&{\revise $2.92\!\pm\!0.04$}
&{\revise $3.23\!\pm\!0.01$}&{\revise $3.16\!\pm\!0.01$}&{\revise $3.48\!\pm\!0.21$}
&{\revise $2.66\!\pm\!0.05$}&{\revise $29.73\!\pm\!0.04$}\\ 
Ours (wo/ Align)   
&{\revise $2.33\!\pm\!0.05$}&{\revise $2.53\!\pm\!0.02$}&{\revise $2.82\!\pm\!0.00$}
&{\revise $3.19\!\pm\!0.00$}&{\revise $3.10\!\pm\!0.00$}&{\revise $2.90\!\pm\!0.21$}
&{\revise $2.53\!\pm\!0.02$}&{\revise $30.04\!\pm\!0.03$}\\
Ours (Align-CR)   
&{\revise $\mathbf{2.18\!\pm\!0.01}$}&{\revise $\mathbf{2.42\!\pm\!0.01}$}&{\revise $\mathbf{2.73\!\pm\!0.03}$}
&{\revise $\mathbf{3.16\!\pm\!0.05}$}&{\revise $\mathbf{3.06\!\pm\!0.04}$}&{\revise $\mathbf{2.62\!\pm\!0.02}$}
&{\revise $\mathbf{2.37\!\pm\!0.01}$}&{\revise $\mathbf{30.54\!\pm\!0.03}$}\\
\hline\hline
\multicolumn{9}{l}{\multirow{2}{*}{(b) {\revise results on SAM and SSIM, to quantify the spectral and structural similarity} }}\\
\multicolumn{9}{c}{}\\\hline
\multirow{2}{*}{} & \multicolumn{6}{c}{per class SAM ($^\circ$) $\downarrow$} 
& \multirow{2}{*}{SAM ($^\circ$) $\downarrow$} & \multirow{2}{*}{SSIM $\uparrow$} \\ \cline{2-7}
&Forest &Rangeland &Agriculture &Urban &Barren &Water & & \\ \hline
McGAN~\citep{enomoto2017filmy}
&{\revise $9.91\!\pm\!0.40$}&{\revise $11.8\!\pm\!0.28$}&{\revise $13.1\!\pm\!0.35$}
&{\revise $14.0\!\pm\!0.67$}&{\revise $13.8\!\pm\!0.68$}&{\revise $15.7\!\pm\!0.01$}
&{\revise $12.3\!\pm\!0.05$}&{\revise $0.783\!\pm\!0.015$}\\
SpA GAN~\citep{pan2020cloud}
&{\revise $8.80\!\pm\!0.27$}&{\revise $8.37\!\pm\!0.24$}&{\revise $8.19\!\pm\!0.30$}
&{\revise $9.82\!\pm\!0.39$}&{\revise $9.19\!\pm\!0.43$}&{\revise $15.4\!\pm\!1.97$}
&{\revise $8.77\!\pm\!0.35$}&{\revise $0.826\!\pm\!0.006$}\\
\hline
SAR-Opt-cGAN~\citep{grohnfeldt2018conditional}
&{\revise $8.26\!\pm\!0.80$}&{\revise $6.94\!\pm\!0.49$}&{\revise $6.86\!\pm\!0.08$}
&{\revise $7.82\!\pm\!0.35$}&{\revise $7.00\!\pm\!0.15$}&{\revise $10.9\!\pm\!0.23$}
&{\revise $7.66\!\pm\!0.44$}&{\revise $0.839\!\pm\!0.021$}\\
DSen2-CR~\citep{meraner2020cloud}
&{\revise $7.10\!\pm\!0.15$}&{\revise $6.55\!\pm\!0.07$}&{\revise $6.42\!\pm\!0.08$}
&{\revise $8.04\!\pm\!0.22$}&{\revise $7.42\!\pm\!0.17$}&{\revise $9.98\!\pm\!0.22$}
&{\revise $6.68\!\pm\!0.08$}&{\revise $0.882\!\pm\!0.002$}\\
GLF-CR~\citep{GLF-CR}  
&{\revise $5.76\!\pm\!0.03$}&{\revise $5.71\!\pm\!0.00$}&{\revise $5.78\!\pm\!0.04$}
&{\revise $7.85\!\pm\!0.02$}&{\revise $7.07\!\pm\!0.02$}&{\revise $9.90\!\pm\!0.11$}
&{\revise $5.65\!\pm\!0.03$}&{\revise $0.906\!\pm\!0.000$}\\
\hline
Ours (wo/ SAR)    
&{\revise $6.59\!\pm\!0.44$}&{\revise $6.07\!\pm\!0.19$}&{\revise $6.24\!\pm\!0.11$}
&{\revise $7.75\!\pm\!0.08$}&{\revise $7.09\!\pm\!0.04$}&{\revise $10.2\!\pm\!0.11$}
&{\revise $6.27\!\pm\!0.27$}&{\revise $0.898\!\pm\!0.005$}\\
Ours (wo/ Align)   
&{\revise $5.93\!\pm\!0.12$}&{\revise $5.71\!\pm\!0.03$}&{\revise $5.70\!\pm\!0.06$}
&{\revise $\mathbf{7.52\!\pm\!0.01}$}&{\revise $\mathbf{6.88\!\pm\!0.03}$}&{\revise $9.67\!\pm\!0.02$}
&{\revise $5.68\!\pm\!0.11$}&{\revise $0.905\!\pm\!0.003$}\\
Ours (Align-CR)   
&{\revise $\mathbf{5.52\!\pm\!0.04}$}&{\revise $\mathbf{5.46\!\pm\!0.09}$}&{\revise $\mathbf{5.55\!\pm\!0.08}$}
&{\revise $7.56\!\pm\!0.16$}&{\revise $6.92\!\pm\!0.17$}&{\revise $\mathbf{9.60\!\pm\!0.30}$}
&{\revise $\mathbf{5.35\!\pm\!0.05}$}&{\revise $\mathbf{0.914\!\pm\!0.001}$}\\
\hline
\end{tabular}
}
\end{table*}

\begin{figure*}[!t]
\centering
\includegraphics[width=1.\linewidth]{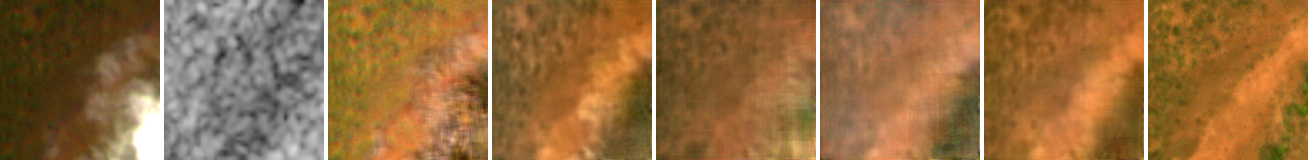}
\\ \vspace{0.1cm}
\includegraphics[width=1.\linewidth]{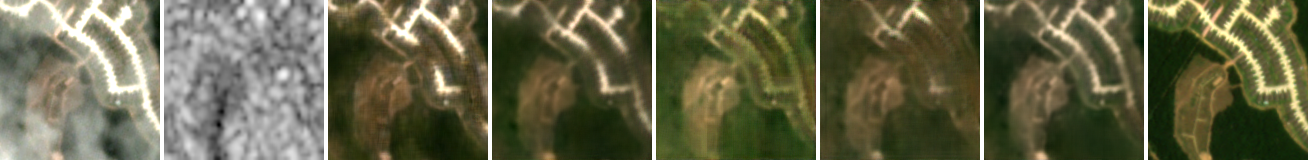}
\\ \vspace{0.1cm}
\includegraphics[width=1.\linewidth]{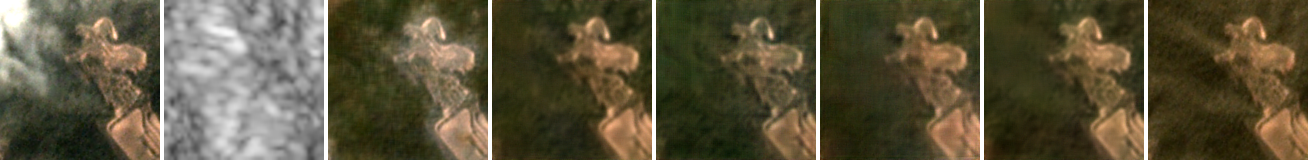}
\\ \vspace{0.1cm}
\includegraphics[width=1.\linewidth]{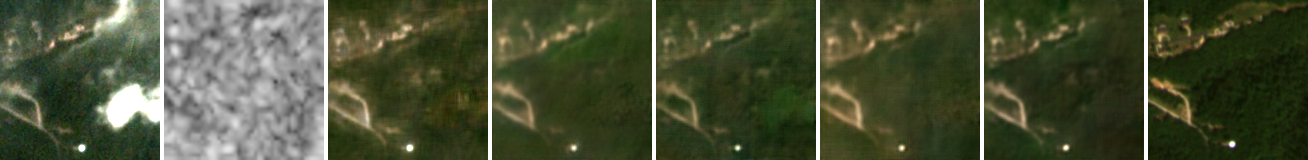}
\\ \vspace{0.1cm}
\includegraphics[width=1.\linewidth]{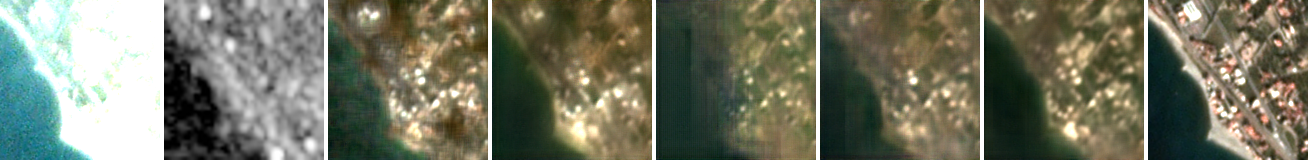}
\\ \vspace{0.1cm}
\includegraphics[width=1.\linewidth]{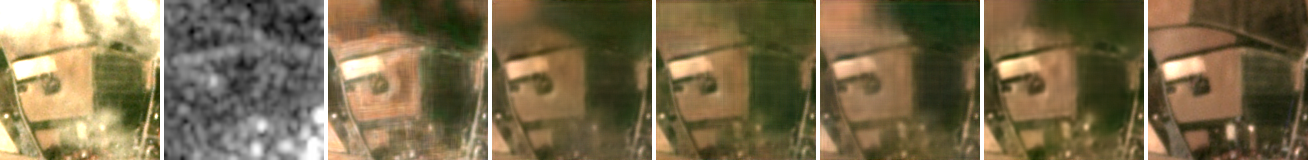}
\\ \vspace{0.1cm}
\includegraphics[width=1.\linewidth]{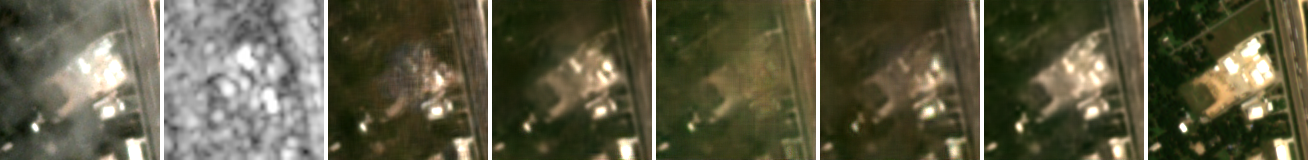}
\\ \vspace{0.1cm}
\includegraphics[width=1.\linewidth]{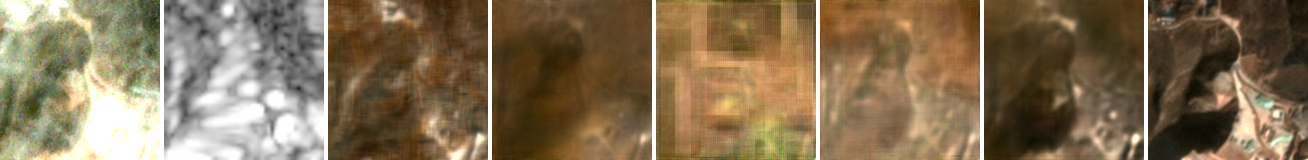}
\\
\includegraphics[width=1.\linewidth]{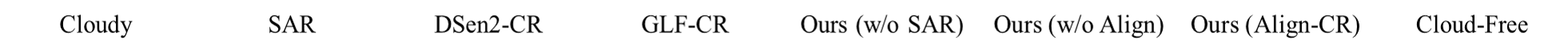}	 
\caption{Qualitative results of visual recovery quality for 8 different samples. The first row shows the cloudy images, the second row shows the SAR images, the third to seventh rows show the results from the DSen2-CR, GLF-CR, w/o SAR, w/o Align and Align-CR models, and the eighth row shows the cloud-free images.}	
\label{fig:visual_result}
\end{figure*}

\begin{figure*}[!t]
\centering
\includegraphics[width=0.85\linewidth]{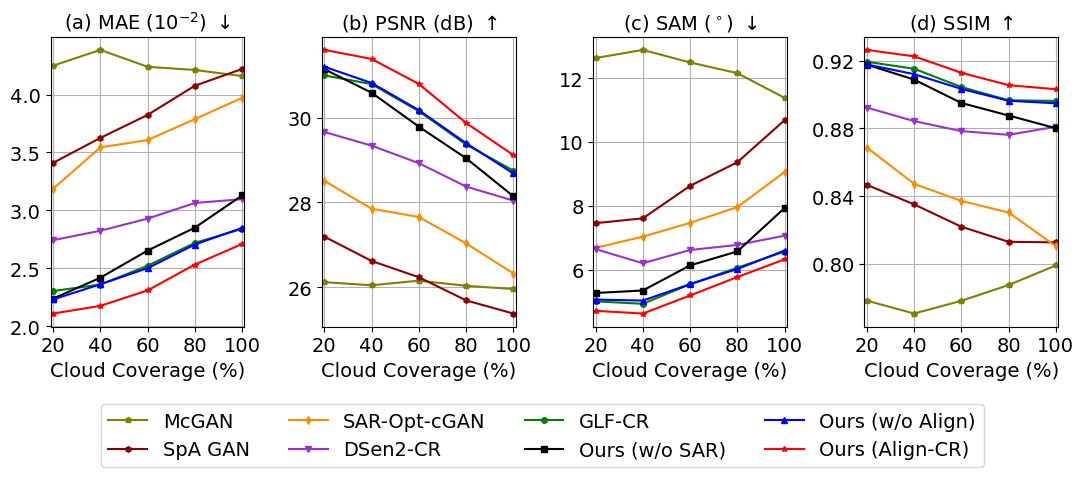}    
\caption{Quantitative results of visual recovery quality over different cloud cover levels in terms of the MAE, PSNR, SAM, and SSIM quality metrics.}	 
\label{fig:cloud_result}
\end{figure*}

\subsection{Evaluation of Visual Recovery Quality}
\label{sec:visual}
To evaluate the quality of the reconstructed images in terms of generating visually pleasing textures, {\revise we start by reporting the Mean Absolute Error (MAE) and Peak Signal-to-Noise Ratio (PSNR) as quantitative measures to assess the reconstruction error. These metrics offer valuable insights into the fidelity of the reconstructed images when compared to their original counterparts. Additionally, we utilize the Spectral Angle Mapper (SAM) and the Structural Similarity Index Measure (SSIM) to further evaluate the quality of the reconstructed images from the perspectives of spectral and structural similarity, which are more closely aligned with human visual perception.}  Notably, with the benefit of the land cover annotations, we disentangle the performance of cloud removal methods over different land cover types by pixel-wise MAE and SAM metrics, as shown in Tab.~\ref{tab:sota}. 
The proposed Align-CR method outperforms the selected state-of-the-art cloud removal methods across all evaluated metrics, demonstrating its superiority in addressing the problem of MMRF-CR. 
It achieves a gain of about 1.68dB compared to DSen2-CR and a gain of about 0.54dB compared to GLF-CR in terms of PSNR.
The McGAN, SpA GAN, and SAR-Opt-cGAN methods, which were developed on relatively small datasets with geographically and topographically homogeneous data, perform poorly in terms of generalizability on our \datasetname{} dataset. 
Their results have relatively limited spectral fidelity, as suggested by the values of the SAM metric in Tab.~\ref{tab:sota}. 
It indicates the need to take the point of global distribution into consideration when creating a practical dataset. 
In Fig.~\ref{fig:visual_result}, we choose 8 scenes to present a qualitative analysis of the reconstructed images.
Our Align-CR method is able to handle various types of clouds, restoring images with more details and fewer artifacts.
Moreover, we compare the results over different land cover types.
We can find that removing clouds over urban land that has highly complex geometrical structures is more challenging than others. 
Urban land in SAR images is characterized by bright-pixel clusters with indistinct boundaries and structures, making it difficult to reconstruct corresponding pixels based on the spectral and texture characteristics of optical images.
Besides, we can find that the results over water do not perform well on SAM but relatively well on MAE. It indicates the challenge of maintaining spectral fidelity over water.

To determine what contributes to the superior performance of the proposed method, we analyze the effectiveness of each component by comparing the proposed method and its variants, i.e., w/o SAR, and w/o Align. 
As shown in Tab.~\ref{tab:sota}, we can observe that the performance of our methods incorporating SAR images is superior to that of w/o SAR over all types of land cover, highlighting the advantages of utilizing SAR images in cloud removal tasks.
Furthermore, as shown in Fig.~\ref{fig:visual_result}, we can find that w/o SAR tends to generate undesirable artifacts for cloud-covered regions due to the lack of ground information. While Align-CR exploits the geometrical information embedded in the SAR images, which can reconstruct the ground object. 
We further compare the cloud removal performance of w/o Align to Align-CR. When the Alignment blocks are removed, the gain of integrating low-resolution SAR information is reduced. 
In regards to the misalignment, such as the junction of different land cover types in the first sample and the river border in the fifth sample in Fig.~\ref{fig:visual_result}, w/o Align tends to generate blurring artifacts, while the reconstructed results of Align-CR have sharper edges. 
It demonstrates the superiority of our proposed Align-CR method in integrating multi-modal and multi-resolution information.

We further assess the reconstruction performance on different cloud cover levels, as shown in Fig.~\ref{fig:cloud_result}. The performance of all methods, except McGAN, decreases roughly as the percentage of cloud cover increases. Among them, the performance of the methods with the benefit of SAR images degrades more slowly than the one without, since the utilization of SAR images can alleviate the decline to some extent. 
Align-CR performs favorably when compared with all baseline methods. When the Alignment blocks are removed, w/o Align behaves very similarly to GLF-CR, while GLF-CR contains more Transformer layers for global fusion and thus performs slightly better in terms of SSIM when more prior information from cloud-free regions is available. Our method Align-CR, which aligns the multi-modal and multi-resolution data during the reconstruction process, can better exploit the power of SAR information. It steadily outperforms w/o Align on all cloud cover levels.

\subsection{Evaluation of Semantic Recovery Quality}
\label{sec:semantic}
\begin{table*}[!t]
\setlength\tabcolsep{3pt}
\caption{The performance of semantic recovery quality over different cloud cover levels.}
\label{tab:ss}
\centering
\resizebox{1.\textwidth}{!}{
\begin{tabular}{lccccccccccccccccc}
\hline
& \multicolumn{2}{c}{0$\sim$20\%} & & \multicolumn{2}{c}{20$\sim$40\%} & 
& \multicolumn{2}{c}{40$\sim$60\%} & & \multicolumn{2}{c}{60$\sim$80\%} & 
& \multicolumn{2}{c}{80$\sim$100\%} & & \multicolumn{2}{c}{Overall}\\ 
\cline{2-3} \cline{5-6} \cline{8-9} \cline{11-12} \cline{14-15} \cline{17-18}
& mIoU & PA  &  & mIoU & PA  &  & mIoU & PA  &  & mIoU & PA  &  & mIoU & PA  &  & mIoU & PA  \\ \hline
Cloud-Free      
& 44.63&62.68&  & 43.66&65.44&  & 41.85&64.40&  
& 40.58&64.16&  & 42.98&66.70&  & 43.24&64.66\\
Cloudy          
& 18.89&34.08&  & 13.26&24.84&  & 10.24&19.27&  
&  8.62&16.21&  &  7.89&12.99&  & 11.72&21.46\\ \hline
McGAN~\citep{enomoto2017filmy} 
&{\revise 14.24}&{\revise 33.95}&  &{\revise 14.79}&{\revise 36.89}&  &{\revise 12.68}&{\revise 32.08}&  
&{\revise 12.30}&{\revise 31.29}&  &{\revise 10.22}&{\revise 32.48}&  &{\revise 13.42}&{\revise 33.29}\\
SpA GAN~\citep{pan2020cloud} 
&{\revise 21.57}&{\revise 39.07}&  &{\revise 19.85}&{\revise 38.36}&  &{\revise 16.54}&{\revise 35.74}&  
&{\revise 13.94}&{\revise 33.30}&  &{\revise 11.46}&{\revise 32.02}&  &{\revise 16.01}&{\revise 35.66}\\
SAR-Opt-cGAN~\citep{grohnfeldt2018conditional}
&{\revise 12.59}&{\revise 30.70}&  &{\revise 10.95}&{\revise 29.38}&  &{\revise 9.96}&{\revise 29.36}&  
&{\revise 8.81}&{\revise 28.56}&  &{\revise 8.30}&{\revise 26.52}&  &{\revise 10.19}&{\revise 28.90}\\
DSen2-CR~\citep{meraner2020cloud}
&{\revise 32.96}&{\revise 43.49}&  &{\revise 29.75}&{\revise 38.90}&  &{\revise 25.96}&{\revise 36.99}&  
&{\revise 22.91}&{\revise 36.05}&  &{\revise 22.91}&{\revise 41.49}&  &{\revise 27.97}&{\revise 39.40}\\
GLF-CR~\citep{GLF-CR}
&{\revise 34.80}&{\revise 48.32}&  &{\revise 32.38}&{\revise 44.96}&  &{\revise 27.27}&{\revise 40.68}&  
&{\revise 23.00}&{\revise 39.33}&  &{\revise 23.64}&{\revise 44.79}&  &{\revise 29.34}&{\revise 43.69}\\
Ours (w/o SAR)  
&{\revise 33.92}&{\revise 47.76}&  &{\revise 27.68}&{\revise 41.49}&  &{\revise 23.94}&{\revise 39.42}&  
&{\revise 19.11}&{\revise 38.11}&  &{\revise 13.50}&{\revise 41.94}&  &{\revise 23.96}&{\revise 41.75}\\
Ours (w/o Align)
&{\revise 35.32}&{\revise 48.94}&  &{\revise 31.70}&{\revise 43.34}&  &{\revise 28.36}&{\revise 41.75}&  
&{\revise 23.80}&{\revise 40.86}&  &{\revise \textbf{24.59}}&{\revise 45.87}&  &{\revise 29.78}&{\revise 44.17}\\
Ours (Align-CR) 
&{\revise \textbf{37.41}}&{\revise \textbf{51.80}}&  &{\revise \textbf{32.66}}&{\revise \textbf{45.32}}&  &{\revise \textbf{29.18}}&{\revise \textbf{43.73}}& 
&{\revise \textbf{24.92}}&{\revise \textbf{42.89}}&  &{\revise 23.29}&{\revise \textbf{46.28}}&  &{\revise \textbf{30.63}}&{\revise \textbf{46.02}}\\
\hline %
\end{tabular}
}
\end{table*}

\begin{figure}[!t]
\centering
\includegraphics[width=1.\linewidth]{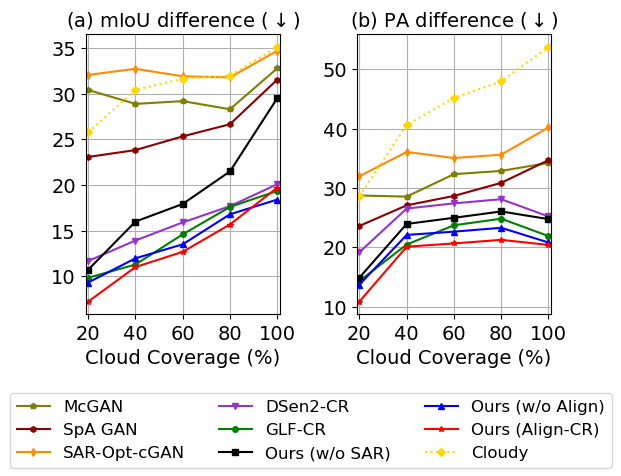}
\caption{The discrepancy between the results obtained using cloud-free images as input and the results obtained using predicted cloud-free images or cloudy images as input.}	
\label{fig:landcover_result}
\end{figure}

\begin{figure*}[!t]
\captionsetup[subfloat]{labelfont={scriptsize}}
\centering
\subfloat[\scriptsize{Cloudy}]{\includegraphics[width=0.135\linewidth]{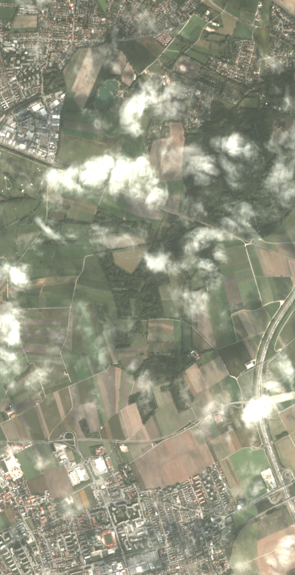}}%
\hspace{0.1mm}
\subfloat[\scriptsize{Cloud-Free}]{\includegraphics[width=0.135\linewidth]{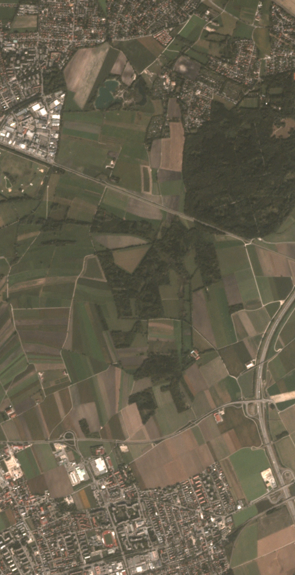}}%
\hspace{0.1mm}
\subfloat[\scriptsize{Align-CR}]{\includegraphics[width=0.135\linewidth]{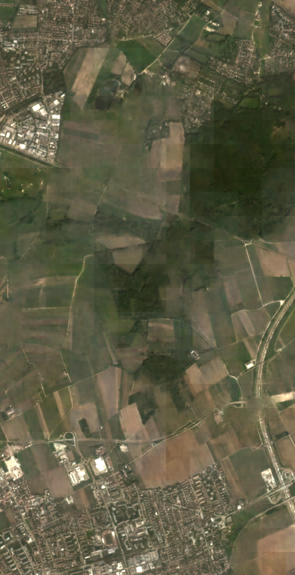}}%
\hspace{0.1mm}
\subfloat[\scriptsize{LC}]{\includegraphics[width=0.135\linewidth]{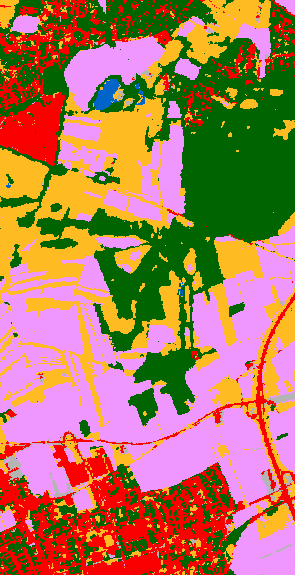}}%
\hspace{0.1mm}
\subfloat[\scriptsize{LC from Cloudy}]{\includegraphics[width=0.135\linewidth]{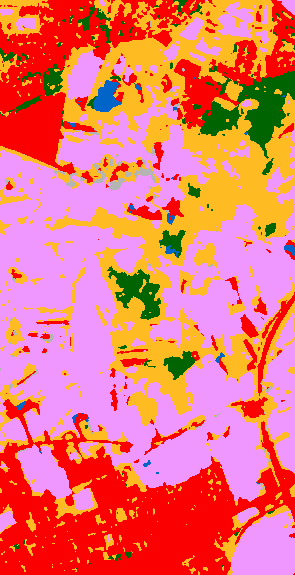}}%
\hspace{0.1mm}
\subfloat[\scriptsize{LC from Cloud-Free}]{\includegraphics[width=0.135\linewidth]{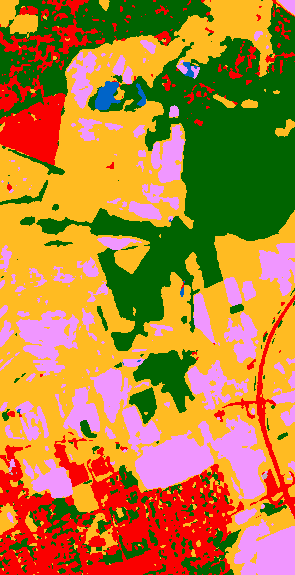}}%
\hspace{0.1mm}
\subfloat[\scriptsize{LC from Align-CR}]{\includegraphics[width=0.135\linewidth]{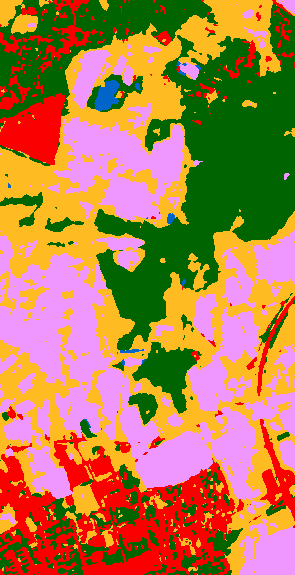}}%
\caption{Visualization of the land cover mapping results. (a) Cloudy image. (b) Cloud-free image. (c) Reconstruction result of Align-CR. (d) Land cover map from WorldCover. (e) Land cover mapping results obtained using a cloudy image as input. (f) Land cover mapping results obtained using a cloud-free image as input. (g) Land cover mapping results obtained using the reconstruction result of Align-CR as input. }
\label{fig:visual_landcover_result}
\end{figure*}

In addition to evaluating the quality of reconstructed images in generating visually pleasing textures, we further evaluate their ability to generate semantically meaningful structures, as semantic information is crucial for future analytical applications.
In this paper, we assess the quality of the recovered semantic information using a well-established land cover semantic segmentation model.
We train a land cover semantic segmentation model using the cloud-free images and associated land cover annotations, using the same data splits adopted for the experiments on cloud removal. The model is based on DeepLabv3plus~\citep{chen2018encoder} with ResNet-50~\citep{he2016deep} as the backbone network. We evaluate the performance of the trained land cover semantic segmentation model in predicting the correct class of each pixel by using cloud-free images, cloudy images, and predicted cloud-free images from baseline models as inputs respectively. 
Ideally, the results with the predicted images as input should be as consistent as possible with the results with the cloud-free images as input, i.e., the closer they are to the results with the cloud-free images as input, the better the corresponding cloud removal method performs in terms of semantic recovery. We report the mean intersection over union (mIoU) and pixel accuracy (PA) over varying levels of cloud cover in Tab.~\ref{tab:ss}, where mIoU can better deal with the class imbalance issue. 
Additionally, we show the discrepancy between the results obtained using cloud-free images as input and the results obtained using predicted cloud-free images or cloudy images as input in Fig.~\ref{fig:landcover_result}.

Clearly, the existence of clouds deteriorates the semantic analysis, and the degradation is more severe when the cloud cover level is higher. The benchmarked cloud removal models can to some extent counteract the degradation, in which Align-CR generally performs the best. Notably, we can observe that the trained land cover semantic segmentation model performs better with the predicted images of McGAN and SpA GAN as input than with the predicted images of SAR-Opt-cGAN as input. While SAR-Opt-cGAN performs better than McGAN and SpA GAN in terms of all visual quality metrics, as shown in Fig.~\ref{fig:cloud_result}. It indicates that the metrics for measuring the visual recovery quality can not adequately reflect the performance of cloud removal methods in terms of semantic recovery. 
We can also find this by comparing the performance of w/o Align and Align-CR. Align-CR steadily outperforms w/o Align on all cloud cover levels in terms of all visual quality metrics. However, it does not perform as well as w/o Align on the images with cloud cover 80\% to 100\% in terms of mIoU.
What's more, we can find that its performance in terms of mIoU is superior to that of w/o Align when more prior information from cloud-free regions is available, since there is more information available for alignment.

In Fig.~\ref{fig:visual_landcover_result}, The land cover mapping results are visualized for comparison. 
We can find that the land cover mapping result obtained using a cloud-free image as input shows the highest consistency with the land cover map from WorldCover. When the image is disturbed by clouds, the corresponding land cover mapping result is significantly negatively impacted. We apply our Align-CR method to reconstruct the image obscured by clouds, as illustrated in Fig.~\ref{fig:visual_landcover_result}(c). The method effectively removes cloud cover and enhances the overall visual quality of the resulting image. By using the reconstructed image as input for predicting land cover maps, we can effectively mitigate the performance degradation issue caused by cloud interference, and significantly improve the accuracy of the mapping result. It shows the effectiveness of our method in both generating visually pleasing textures and generating semantically meaningful structures. 

\section{Discussion}
\label{sec:discussion}

\noindent{\bf Semantic Considerations in Cloud Removal Beyond Visual Evaluation.}
Due to the limited availability of pixel-wise semantic annotations provided along with the cloud removal datasets, most existing studies rely solely on metrics that evaluate the visual similarity between two images when quantifying the effectiveness of cloud removal methods. Towards performance gains in visual metrics, it is common to use loss functions that are constructed based on these metrics to guide the training of cloud removal models, e.g., the L1 loss function computes the mean absolute error~\citep{GLF-CR}. It will motivate the predicted cloud-free image to move towards over-smoothness and lead to the potential loss of semantic information. As validated in Sec.~\ref{sec:semantic}, the current visual metrics cannot adequately evaluate the quality of recovered semantic information. Although our method achieves the best results in terms of semantic recovery, there remains a significant discrepancy between the predicted cloud-free images and their corresponding real cloud-free images. Consequently, it is crucial to develop loss functions that can effectively guide the recovery of semantic information. Future research could focus on the coupling between cloud removal and downstream tasks and leverage this relationship to design a loss function that balances both the semantic context and image details, thus providing more meaningful and practically useful guidance for the reconstruction process.

\begin{figure*}[!t]
\centering
\includegraphics[width=1.\linewidth]{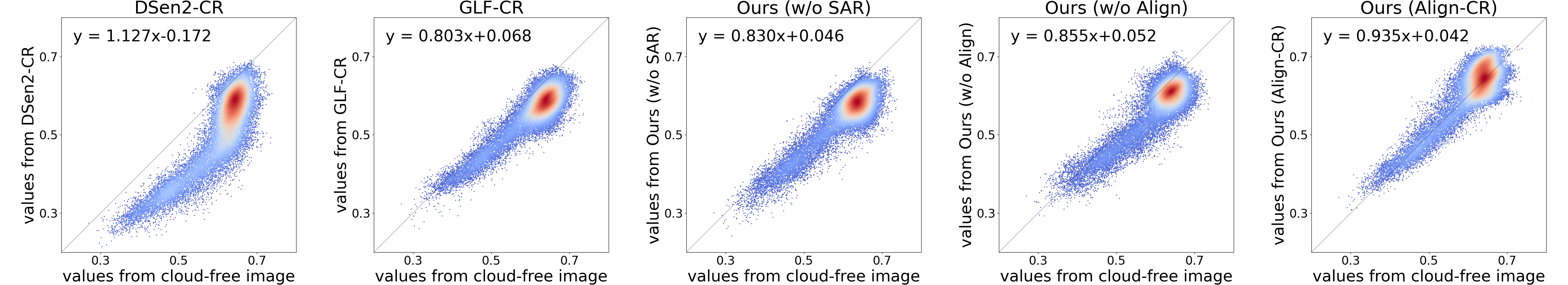}\\
\vspace{0.2cm}
\includegraphics[width=1.\linewidth]{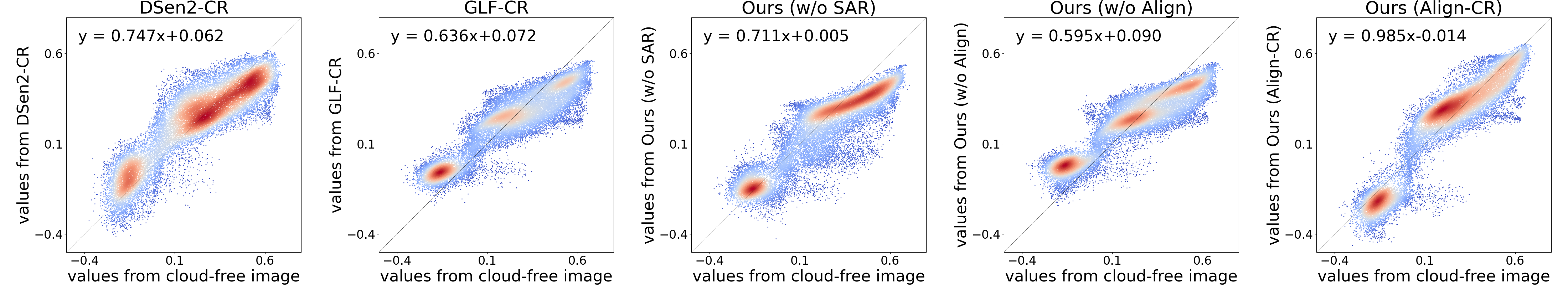}
\caption{Scatter plots comparing NDVI values derived from cloud-free images with those from reconstructed images.}	
\label{fig:ndvi_scatter}
\end{figure*}

\begin{table}[!t]
\caption{Quantitative comparisons of the NDVI values derived from cloud-free images with those from reconstructed images.}
\label{tab:ndvi}
\centering
\begin{tabular}{lcc} \hline
                                                 & RMSE $\downarrow$   &mIoU $\uparrow$  \\ \hline
McGAN~\citep{enomoto2017filmy}                   & 0.1878 &16.20 \\
SpA GAN~\citep{pan2020cloud}                     & 0.2417 &23.28 \\ 
SAR-Opt-cGAN~\citep{grohnfeldt2018conditional}   & 0.1126 &22.37 \\ 
DSen2-CR~\citep{meraner2020cloud}                & 0.1015 &30.78 \\ 
GLF-CR~\citep{GLF-CR}                            & 0.0845 &32.06 \\ \hline
Ours (w/o SAR)                                   & 0.1003 &28.17 \\ 
Ours (w/o Align)                                 & 0.0844 &32.14 \\ 
Ours (Align-CR)                                  & \textbf{0.0802} &\textbf{33.77} \\ \hline
\end{tabular}
\end{table}

{\revise 
\noindent{\bf NDVI-Based Evaluation} 
In addition to the evaluations on visual and semantic recovery quality detailed in Sec.~\ref{sec:visual} and Sec.~\ref{sec:semantic}, we also harness the Normalized Difference Vegetation Index (NDVI), a widely used remote sensing index that offers invaluable insights into vegetation land cover detection and interpretation, to enrich our analysis.
The NDVI calculation hinges on the spectral reflectance differences between the Near Infrared and red bands, with values spanning from -1 to 1. A higher value correlates with increased vegetation density. 
Typically, the values can be categorized into five levels: values less than $-0.1$ depict areas like ground cloud, water, and snow cover; values between $-0.1$ and $0.1$ indicate areas of rock and soil; values between $0.1$ and $0.4$ signify sparsely vegetated areas; values between $0.4$ and $0.8$ denote densely vegetated areas; and values greater than or equal to $0.8$ characterize super-densely vegetated areas~\citep{zhang2022two}. 
\begin{figure}[!t]
\centering
\includegraphics[width=1.\linewidth]{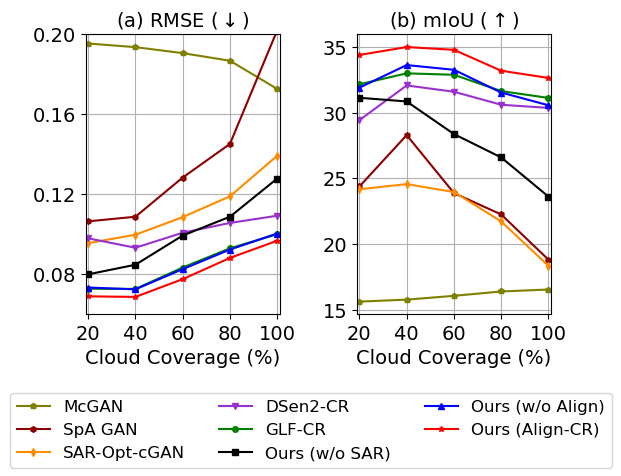}
\caption{Comparison of the NDVI values over different cloud cover levels.}	
\label{fig:ndvi_cloud}
\end{figure}
By comparing the NDVI values extracted from the recovered images to those from the cloud-free images, we are able to evaluate the methods' capability in preserving the distinctive land cover characteristics within the recovered images. 
In this study, we utilize the Root Mean Square Error (RMSE) to quantify the discrepancies between NDVI values extracted from the recovered images and those from the cloud-free images. Additionally, by taking the vegetable classification results derived from the cloud-free NDVI values as the reference, we employ mIoU to evaluate the precision of the classification based on the de-clouded NDVI values.
The corresponding results are presented in Tab.~\ref{tab:ndvi}. 
Remarkably, the Align-CR method exhibits a higher overall accuracy when compared to other approaches.
To offer a more intuitive representation, we employ scatter plots that depict the NDVI values. 
Specifically, in Fig.~\ref{fig:ndvi_scatter}, we showcase the NDVI values for two distinct samples from Fig.~\ref{fig:visual_result}: the third sample covering both densely and sparsely vegetated regions, and the fifth sample highlighting water bodies and urban terrains.
We can find that the NDVI values derived using the Align-CR method more closely align with those from cloud-free images, indicating the superior capability of our method in preserving land cover characteristics. Furthermore, the points representing water bodies, urban areas, and vegetated zones exhibit heightened concentration, reinforcing the effectiveness of Align-CR in capturing distinct NDVI values across various land cover types.
Delving further, we illustrate the comparison of NDVI values over different cloud cover levels, as shown in Fig.~\ref{fig:ndvi_cloud}. 
The Align-CR method consistently outperforms competitors across all cloud cover levels in terms of both RMSE and mIoU metrics.
It is worth noting that the trend in RMSE correlates with cloud cover levels in a manner analogous to that observed for SAM.
This parallelism arises because the NDVI is computed based on the spectral correlation of the reconstructed images, which closely aligns with the SAM. As for mIoU, it leans more towards a semantic understanding of the scene, offering a perspective distinct from the aforementioned visual quality metrics.
}

{\revise
\noindent{\bf Model and Computational Complexity.}
In Table~\ref{tab:flops}, we present a comparison detailing the number of model parameters, floating-point operations (FLOPs), and the inference time when evaluated on an NVIDIA GeForce RTX 3090 GPU with a batch size of 1.  
Notably, while Align-CR exhibits higher model complexity, it significantly reduces computational complexity compared to DSen2-CR. 
Furthermore, Align-CR achieves a slightly faster inference speed than that of GLF-CR. 
This improvement can be attributed to our strategic reduction in the complexity of the SGCI block, as explained in Sec.~\ref{sec:fuse}.
}

\begin{table}[!t]
\caption{Comparison of the number of parameters, FLOPs, and inference time for various models.}
\label{tab:flops}
\centering
\begin{tabular}{lccc} \hline
                                    & \#Params & \#FLOPs  & Inference Time   \\ \hline
DSen2-CR~\citep{meraner2020cloud}   & 18.91M    & 483.77G  & 228.7ms          \\ 
GLF-CR~\citep{GLF-CR}               & 14.68M    & 95.13G   & 318.9ms          \\ 
Align-CR                            & 43.53M    & 174.94G  & 309.4ms          \\ \hline
\end{tabular}
\end{table}

In future research, we will integrate the fusion of multi-temporal, and multi-modal multi-resolution data for high-resolution cloud removal, aiming to develop more advanced methods that leverage the complementary information provided by different data sources at different times to enhance the effectiveness of remote sensing analysis under cloudy conditions.

\section{Conclusion}
\label{sec:conclusion}
We propose \datasetname{} -- an open-source multi-modal and multi-resolution dataset designed specifically for cloud removal in high-resolution optical remote sensing imagery. With this effort, we aim to encourage the development of innovative cloud removal approaches that leverage the integration of multi-modal and multi-resolution information.
To address the challenge of cloud removal in such datasets, we propose a novel method called Align-CR. This method utilizes an implicit alignment of feature maps during the reconstruction process to compensate for misalignment between multi-modal and multi-resolution data. Our experimental results demonstrate the effectiveness and superiority of the Align-CR method compared to existing representative cloud removal methods, both in terms of visual recovery quality and semantic recovery quality.



\section*{Acknowledgement}
Source of the planet data used in the publication: Planet Labs Inc.




\bibliographystyle{IEEEtran}
\bibliography{refs}

\vfill

\end{document}